# COMTEST Project: A Complete Modular Test Stand for Human and Humanoid Posture Control and Balance

Vittorio Lippi, Thomas Mergner, Thomas Seel, Christoph Maurer

*Abstract—* This work presents a system to benchmark humanoid posture control and balance performances under perturbed conditions. The specific benchmarking scenario consists, for example, of balancing upright stance while performing voluntary movements on moving surfaces. The system includes a motion platform used to provide the perturbation, an innovative body-tracking system suitable for robots, humans and exoskeletons, control software and a set of predefined perturbations, a humanoid robot used to test algorithms, and analysis software providing state of the art data analysis used to provide quantitative measures of performance. In order to provide versatility, the design of the system is oriented to modularity: all its components can be replaced or extended according to experimental needs, adding additional perturbation profiles, new evaluation principles, and alternative tracking systems. It will be possible to use the system with different kinds of robots and exoskeletons as well as for human experiments aimed at gaining insights into human balance capabilities.

## I. INTRODUCTION

### A. Background and motivation

Currently, the humanoid robotics community is aiming to introduce quantitative measures of performance [1]. Such measures are envisaged to provide a tool for both the evaluation and the improvement of robots. Posture control and balance present key challenges for the use of humanoids in a real-world scenario, as shown by frequent falls of robots when interacting with a complex environment such as the one proposed by the DARPA challenge [2,3]. In this work, we describe a project that takes inspiration from the techniques used to evaluate human posture control [4] and human-inspired robot control [5]. The motivation is founded both in the current superiority of human posture control that can represent a target for humanoids [1,6,7] and the possibility of using the experience gained through several decades of human experiments in studies such as [8] and [9].

At the state of the art, the perturbations used to test robot balance control usually consist of some form of instability or externally produced movement of the support

This work is supported by the project EUROBENCH (European Robotic Framework for Bipedal Locomotion Benchmarking, www.eurobench2020.eu) funded by H2020 Topic ICT 27-2017 under grant agreement number 779963.

Vittorio Lippi and Thomas Seel are with Technische Universität Berlin, Fachgebiet Regelungssysteme (Control Systems Group). Einsteinufer 17, Berlin, Germany

Thomas Mergner and Christoph Maurer are with the University Clinic of Freiburg, Neurology, Freiburg, Germany

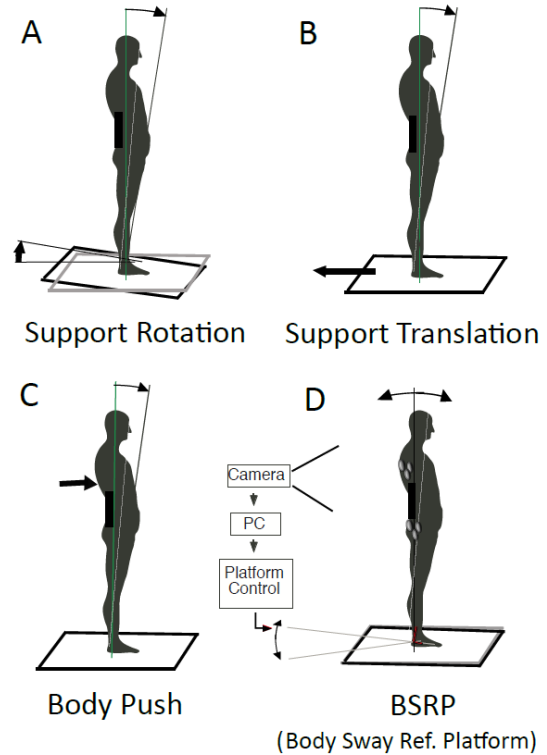

*Figure 1 Suggested posture control disturbance scenarios (inspired by human studies). Examples refer to single inverted pendulum scenarios that challenge the balancing of biped standing in the sagittal body plane (with moderate stimuli mainly around the ankle joints). (A) Support surface rotation about the ankle joint axis [28]. (B) Support surface translation. (C) Contact force stimulus (applied as pull on a body harness using cable winches). (D) "Body sway referenced platform" (tests selectively vestibular system).*

surface (like in [10]) or collision with a swinging weight (see for example [11]). Such tests can assess qualitatively the ability of the robot to balance in some scenarios, but do not provide a quantitative measure or a repeatable benchmark to evaluate posture control.

Posture control refers here to the ability to maintain the body center of mass (CoM) above the base of support, compensating external disturbances. Furthermore, posture control is active during voluntary movements and represents a prerequisite for gait. Referring to the schema presented in [6], the proposed system consists of a testbed that allows for the evaluation of the humanoid robots within the experimental set-up as presented in the human posture control literature (e.g. [8,9,12]). Specifically, the perturbations consist of surface motions and the measured responses are the sways of the body segments [4, 5].

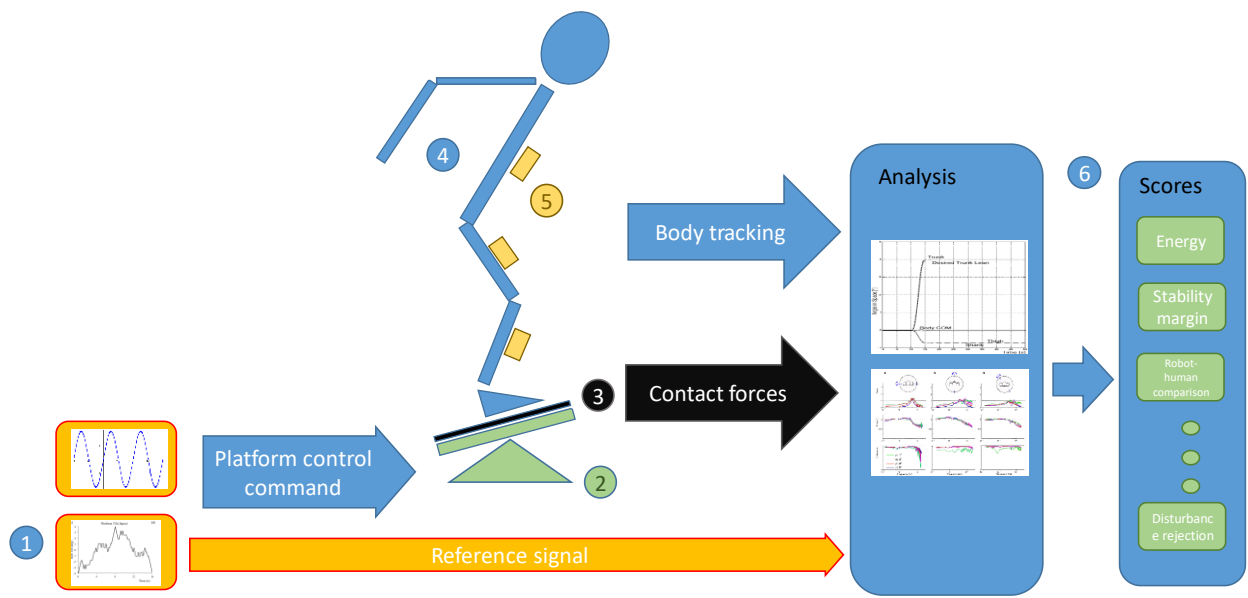

*Figure 2- System architecture. The schema shows the components of the complete system and their mutual relationships. Specifically, (1) is the collection of external stimuli used to control the platform and as an input for the analysis tools, (2) is the platform, that can rotate and translate, (3) is a force plate, (4) the test robotic platform, (5) the magnetometerless wearable body tracking system, (6) the analysis software, producing elaborated data interpretations (e.g. frequency response functions) and scores measuring the performance.*

The perturbations device consists of a moving support surface on which the robot stands. It is designed to produce conditions that reproduce the interaction between the body and the environment. The evaluation is based on measured variables (body sway, ground reaction forces) so that the same experimental set-up can be repeated, and the same quantitative criteria can be applied to robots. The evaluation comprises criteria for both assessing the human likeliness of the behavior as well as efficiency and stability measures that are relevant for disturbance rejection.

### B. System Overview

The scenarios will cover four disturbances: (i) support surface tilt, (ii) the field force gravity, (iii) support surface linear acceleration and (iv) external contact force (compare Fig. 1). Gravity is considered an external disturbance although it is not produced actively by the system. This comes from the idea that it requires specific compensation based on sensory input. The support surface tilt can also be controlled to be equal to the body sway with respect to the gravitational vertical. Such body sway referenced platform control (Fig. 1D) imposes a constant ankle joint angle of 90°, which prevents the robot from using ankle joint encoder information. The choice of a moving support surface as a tool to explore the functionality of human and humanoid posture aims to be *simple*, because the disturbance is then a well-defined input for the system (human or robot), to be *complete* because the interaction of the human body with the external environment can be described by 4 basic disturbances [4] that can be produced with this set-up, to be *oriented to the study of the sensor fusion* because, compensating the effect of a moving surface requires the integration of body link motion perception (encoders for robots, proprioception for humans) and of position in space (IMU for robots, vestibular system for humans). In general, measuring the abilities required to stand the 4 disturbances allows us to decompose the global performance into several measurable and meaningful components, e.g. rejection of a specific disturbance or the ability to perform voluntary movements in the presence of various disturbances scenarios. The examples shown in Fig.1 refer to a single inverted pendulum (SIP) scenario. This is motivated by the fact that, in presence of small stimuli, humans tend to respond primarily in the ankle joints only (yielding a SIP, single inverted pendulum, scenario, where it is enough to evaluate the SIP angle in space). With larger stimuli (i.e. evoking body sway larger than 3°) the response tends to include the hip joints, resulting in a double inverted pendulum, DIP, scenario. The system allows for the analysis of both the cases shown in [13]. Disturbance compensation can be also studied in superposition with self-generated disturbances from voluntary movements associated with motor tasks, e.g. superposition of active hip bending movements during external perturbation in the form of support surface tilts [12].

## II. SYSTEM ARCHITECTURE

In detail, the system consists of (i) datasets of predefined perturbations that can be used to test the robot/subject, (ii) a 6 DoF moving support surface, (iii) a body tracking system, (iv) a humanoid platform and (v) an analysis system. This set-up allows experimenters to apply state of the art protocols for posture control analysis e.g. [4,5,8,13]. Benchmarking metrics and algorithms will be designed to allow automatic calculation of performance scores based on recorded data. An overview of the system is shown in Fig. 2. The design of the proposed solution addresses the problem from a general point of view. The structure of the proposed testbed is modular, all the components are suitable to be used together, alone or integrated into new experimental conditions.

*A. Moving Support Surface*

The moving support platform has a size of 1.5 x 1.5 m, it is capable of producing an impulsive translation, impulsive tilt, pseudorandom tilt/translation. The specifications in terms of precision, speed and payload have been defined in order to reproduce the experimental setups shown in [4,5], e.g. PRTS signal with sudden speed variations [4,5,8], foam rubber-like tilting, to mimic to the body-sway-referenced platform (BSRP) tilt of [4], and simulations of a passive seesaw/rocker-board, involving the force platform, and a slippery-like surface (zero shear forces). In particular, the BSRP (Fig. 1 D) is a control modality where the platform tilt is set to be equal to body CoM sway. In a single inverted pendulum scenario, this imposes a fixed ankle position and removes the possibility to base the balance solely on proprioceptive input (encoders). The support platform is based on the 6 DOF device *PS-6TM-150* from *Motionsystems* (www.motionsystems.eu).

*B. Tracking System*

We design the tracking system such that it can be used in humanoid robots as well as in humans with and without exoskeletons. This facilitates the motion analysis to be carried out based on the same measured variables throughout all tests and assures a high level of comparability.

Previously, posture control experiments have been performed with optical tracking e.g. [9,13] or exploiting the same internal sensors used for the control e.g. [11, 14,15] or a mixture of the two solutions e.g. [16]. While optical tracking systems are expensive and limited by line-of-sight restrictions, inertial measurement systems typically require extensive calibration protocols as well as a homogeneous magnetic field [18]. However, these challenges can be addressed using recent advancements in the development of inertial sensor fusion [19,20,32].

The proposed system consists of wearable IMUs that are attached to the body segments as illustrated in Fig. 2 and a set of sensor fusion methods that achieve accurate real-time motion tracking in a plug-and-play manner [21,22,23,33].

Inertial tracking systems that use magnetometers can determine the complete orientation, i.e. the attitude and heading, of an object with respect to an inertial reference frame or another object that is equipped with an IMU. However, this only works well in a homogeneous magnetic field far from ferromagnetic material and electric devices. We therefore employ 6-axis inertial sensor fusion, i.e. we only use accelerometer and gyroscope readings, and determine the heading information by exploitation of kinematic constraints [19–22].

The data analysis is carried out in two layers: one highly accurate offline analysis and an online data processing layer that provides measurements for real-time operations and control such as for body-sway-referenced support surface tilts [6] and may provide position feedback to human subjects.

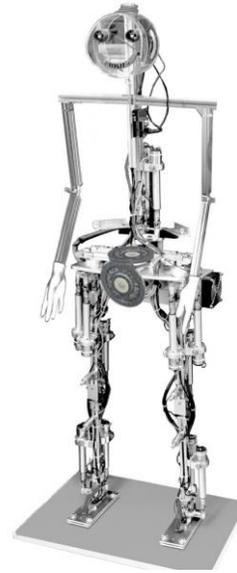

*Figure 3 – The 'Lucy' humanoid robot. Notice the added weight on the hip (two 1kg iron disks)*

*C. Test Humanoid*

A robotic platform based on the open-source humanoid *Poppy* [17] will be included in the system. It is thought to address a large number of potential users represented by the developers of posture control algorithms currently using simulations and to provide a simple, stand-alone, standard platform to compare different control algorithms. A main modification of the original robot endeavor aims to allow for a torque-controlled actuation [12], a prerequisite common for most posture control algorithms at the state of the art. The robot design and control will be released as an open-source project as well.

*D. Analysis Software*

The main aim of the test bench will be evaluating the performance without actually reaching failure in balance control. The analysis of frequency response functions and the characterization of non-linearity shown in [5,9,12] provide descriptive features, on which basis it is possible to compute the desired metrics, as they are designed during the development of the project on the basis of experiments and received feedback. The performance can be measured in terms of torque or energy applied, with dynamic performance indexes (e.g. overshoot and settling time in response to transient stimuli or voluntary movements). Also, periodic stimuli can be used as will be shown in the example in Section III. The analysis and control software will be developed in Python/ROS. The data used in posture control analysis mainly consist of kinematic variables (i.e. body segments sway, estimated CoM sway, and platform position and orientation). Such values can be obtained both with the provided tracking system or with other systems considered suitable for the specific experiment, with kinematic variables allowing for the use of legacy data. The system is envisaged to be integrated into a facility where optical tracking systems such as *Vicon* and *Optotrack* will be available as well as integrated technologies such *as wearHEALTH* system [29,30].

*E. Human Database*

For quantitative benchmarking of the human-like postural features described above, human experimental data serve as the references [26]. In particular, the human database will be used both to compare human and robot performance, applying the same metrics to humans and robots, and to define measures of human likeliness, i.e. measuring differences between humans and robots based on relevant features such as frequency response functions.

### III. EXAMPLE ANALYSIS

In order to provide an example of how the data analysis pipeline of the benchmarking experiment works, a preliminary experiment is reported. The humanoid robot *Lucy*, which uses the disturbance estimation and compensation, DEC, control method [5] with distributed (yet hierarchically interconnected) controllers was used to demonstrate the effects on balancing of lifting an external weight (8.25% of robot weight) in the static phase, while the support surface was concurrently tilted (here sinusoidally) in the body sagittal plane. The humanoid is 1.5 m tall and it weighs approximately 16.5 kg. The body consists of the trunk, pelvis, the thigh, shank and foot segments. The total of 14 DoF comprises 2 DoF per ankle joint, 1 per knee, 3 per hip, and 2 DoF for the trunk. Each degree of freedom is equipped with an encoder and a torque sensor. Fixed to the robot's upper body segment, a bio-inspired vestibular system [9] estimates upper body angular velocity and angle with respect to the gravitational vertical and linear head acceleration. The joints are actuated by force-controlled DC electric motors. Annotating the challenged ability specifically to the primarily relevant DEC mechanism (here for external force compensation) can be achieved by performing parameter estimations of the controller and comparing the perturbed with the unperturbed case.

The robot was tested with a sinusoidal support surface tilt of 2° at 0.05 Hz, while commanded to stand upright. The system was tested in the original set up from [5] and with an additional mass that was not taken into account by the model parameters. As can be noted from Fig. 3, the placement of the iron disks was asymmetric (as if a point mass of 2 Kg was placed at the pelvis at 0.15 m above the ankle joints in an eccentricity with respect to the body plane through the ankle and hip joints. It produced the effect which an external forward pull (contact) force would exert on the pelvis. We conceive it as an elegant way to circumvent the need to involve a ‚pusher' or ‚puller' device for evaluating the response to contact forces (compare Fig. 1C). As can be seen in Fig. 5, the body sway remained here approximately centered around zero (representing the aforementioned body plane), indicating that the pull force was well compensated (in this case by the contact force DEC mechanism; compare [31],

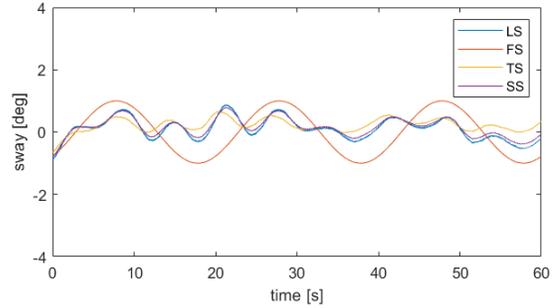

Figure 4- Robot body sway in the sagittal plane shown for each body segment. FS = foot in space corresponds to the rotation of the support surface imposed by the testing device. The other segments are addressed as SS= shank in space, LS= leg in space, TS = trunk in space

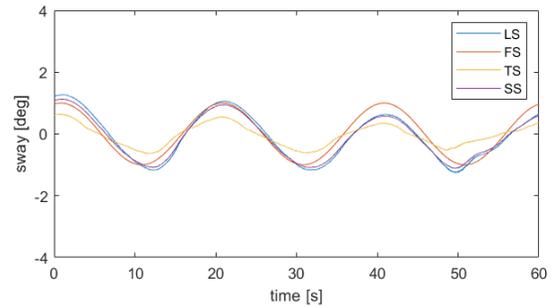

Figure 5- Body segments sway responses to the same stimulus shown in Fig. 3 with the addition that now a non-modeled 2 kg mass has been added to the hip (see Fig. 3). The task of the envisaged analysis tools would be to identify the control mechanisms challenged by the task.

[5]). The use of such simple experiments may contribute to one goal of the *Eurobench* project: the comparison between the performance of different robots in a standardized way.

Adding the weight suddenly may be used as an additional test to dynamically characterize the external force compensation. However, more meaningful results can be obtained using standardized pull (or other) stimuli in the form of a pseudo-random ternary sequence that allows characterizing low-, mid- and high-frequency response features in terms of frequency response functions (FRFs) with gain, phase and coherence (see [4]). In order to compare the performances in both the cases the body sway produced in response to the stimulus has been analyzed and some examples of measures were computed. The stimulus is relatively small with respect to the capabilities of the humanoid, this makes the test safe even considering the effect of the non-modeled additional weight.

The *support surface tilt* disturbance is expressed in terms of angular sway in time under the assumption that the foot of the robot stays in contact with the support, and hence it is addressed as FS = foot in space angle in the Figs. 4 and 5 (note that he actuation of the tilting platform produced the torque required to follow the desired tilt trajectory notwithstanding the weight force and the active torque produced by the robot). In order to compare the effect of external disturbances on robots characterized by different sizes and mass distributions the applied torques can be normalized by ***mgh***, where ***m*** is the mass of the robot, ***g*** the gravity constant and ***h*** the height of the CoM as explained in [13].

IV. RESULTS

The response of the robot in the two conditions is shown in Fig. 4. (robot in nominal conditions) and Fig. 5 (robot with an additional weight on the hip, as shown in Fig. 3). The addition of a non-modeled mass produced a deterioration of the performance manifested in a larger sway of the body segments. In order to directly compare the two different performances, it would be useful to express them through a scalar score.

Three examples are proposed:

| Score | No additional mass | With additional mass |
|---|---|---|
| *Gain* | 0.15 | 0.57 |
| *Phase lag* | -0.35 rad | 0.00 rad |
| *Power* | 1.38 $10^{-5}$ rad$^2$/s | 5.66 $10^{-5}$ rad$^2$/s |

Specifically, *Gain* and *Phase-lag* refer to the gain and phase between the input support surface sway *u(t)* and the sway of the CoM *α(t)*, computed on the basis of the body segment sways shown in Fig. 4 and Fig. 5. They provide an estimate of the sensitivity of the system with respect to the input. The indexes are computed considering the sample cross-correlation between the input and the CoM sway

From the numerical point of view $\alpha$ and $u$ are represented as vectors of samples of finite size *n* and hence $X(t)$ is a vector of *2n-1* samples. The transient response of the system is typically not used for the analysis when working with data obtained with periodic stimuli. This means that several periods in the beginning are discarded and sampling of $\alpha$ and $u$ is started when the system is considered to be in steady-state regime.

Specifically, *Gain, G* is defined as

$$G = \frac{\max_{t}\|X(t)\|}{\sum_{t=-\infty}^{\infty} u^2(t)} \quad (2)$$

and *Phase φ* is

$$\varphi = 2\pi \frac{\operatorname{argmax}_{t}\|X(t)\|}{Tf} \quad (3),$$

where *T* is the period of the input signal and *f* is the sampling frequency (in this discrete case *t* is an index addressing a sample).

The responses in Fig. 4 are not sinusoidal because of the nonlinear nature of the system (owing mainly to bio-inspired thresholds of the disturbance estimates). In order to quantify also the sway produced on other frequencies than the one provided by the input, the score *Power* is computed as the integral of the squared CoM sway over a period, in the sampled version:

$$P = \frac{1}{N} \sum_{t=1}^{N} \alpha(t)^2 \quad (4).$$

Eq. 4 is the typical formula for the power of a signal. In the context of humanoid benchmarking, one could consider the actual mechanical power produced by the actuators (i.e. the product of joint torque and joint speed) like in [27]. In the general case, however, the actual power consumption for a humanoid depends strongly on the actuators and cannot be easily computed from externally measurable variables.

In this simple example, the reason for the performance difference is known - it is the weight that has been artificially added to the body. The *Gain* and *Power* values indicate that in the case without additional mass the robot is performing better (i.e. it shows relatively little sway in response to the external disturbance). The *phase lag* of almost zero in the case with additional weight means that the stimulus and the produced sway are coherent, reflecting that the disturbance has a stronger effect on the output (the robot sways almost with the support base, i.e. its compensation of the disturbing tilt is now clearly smaller). This may provide a hint for the experimenter suggesting that the compensation may profit by an additional mechanism that measures the body weight and adjusts the ***mgh*** value in the corresponding controller parameters (in this case thereby overcoming the mismatch between the nominal parameters and current actual body mass).

V. CONCLUSION AND FUTURE WORK

A major innovation provided in the proposed benchmarking framework is evident in the above example: Relatively small stimuli may suffice to achieve standardization and repeatability of experimental tests of postural and stance stability, this without the need to provoke fall and possibly damage to the robot (or to the human in an exoskeleton). Furthermore, the approach allows combining different external stimuli and to use more than one tool for analysis, and to select among several tools those best suited for a specific experimental question or framework. The system developed for the Eurobench consortium will incorporate a testbed ready for posture control experiments, tested software routines for the analyses, and human datasets as a reference for comparison.

The system is tailored to meet criteria that were developed in the past two decades to study human standing balance quantitatively, such that the user can refer to a standardized database of balancing performance. Using this reference *per se* already aims to evaluate robot performance *without using fall as a criterion*. The modular design of hard and software allows for easy integration of new or additional analyses and testing techniques.

The proposed setup is focused on the analysis of posture and balance control standing, but the task of walking on moving surfaces may in the future also be analyzed using the methods proposed in this framework. For example, a testbed for stepping in place (SIP) balancing tests is currently being developed in Freiburg after Peterka recently showed that the balancing control test described in [8] can, with restrictions, also be applied to the SIP paradigm. Principally, our effort aims to integrate several evaluation indexes that allow for further future developments. The 6-DoF platform used will allow superposition of support surface movements in different planes (e.g. in sagittal and frontal body planes, or tilt combined with horizontal translation, or combined horizontal and vertical motions). The analysis of perturbing three-dimensional motion perturbation may require the development of corresponding new analysis tools, as is also true when prediction and learning are possibly included in the future. Conceivably, this extension will require to include in the modeling of the posture control mechanisms also model-based approaches (reflecting in human brain functions higher order mechanisms than so far considered, for example, in the DEC model).

## VI. ACKNOWLEDGMENT